\documentclass[11pt]{article}
\usepackage{graphicx}
\usepackage{longtable,lscape}
\usepackage{amsfonts}
\usepackage{float}
\usepackage{url}

\newcommand{\captionfonts}{\footnotesize}
\makeatletter  
\long\def\@makecaption#1#2{%
  \vskip\abovecaptionskip
  \sbox\@tempboxa{{\captionfonts #1: #2}}%
  \ifdim \wd\@tempboxa >\hsize
    {\captionfonts #1: #2\par}
  \else
    \hbox to\hsize{\hfil\box\@tempboxa\hfil}%
  \fi
  \vskip\belowcaptionskip}
\makeatother 
\begin{document}
\title{Quantum Structure in Cognition: Fundamentals and Applications}
\author{Diederik Aerts$^1$, Liane Gabora$^2$, Sandro Sozzo$^1$ and Tomas Veloz$^2$ \vspace{0.4 cm} \\ 
        \normalsize\itshape
        $^1$ Center Leo Apostel for Interdisciplinary Studies \\
        \normalsize\itshape
        Brussels Free University, Pleinlaan 2, 1050 Brussels, 
       Belgium \\
        \normalsize
        E-Mails: \url{diraerts@vub.ac.be,ssozzo@vub.ac.be}
        \vspace{0.2 cm} \\
        \normalsize\itshape
        $^2$ Department of Psychology and Computer Science \\
        \normalsize\itshape
       University of British Columbia, Okanagan Campus, \\
        \normalsize\itshape
        3333 University Way, Kelowna, Canada \\
        \normalsize
        E-Mails: \url{liane.gabora@ubc.ca, tomas.veloz@ubc.ca}
        }
\date{}
\maketitle              
\begin{abstract} 
\noindent 
Experiments in cognitive science and decision theory show that the ways in which people combine concepts and make decisions cannot be described by classical logic and probability theory.
This has serious implications for applied disciplines such as information retrieval, artificial intelligence and robotics. Inspired by a mathematical formalism that generalizes quantum mechanics the authors have constructed a contextual framework for both concept representation and decision making, together with quantum models that are in strong alignment with experimental data. The results can be interpreted by assuming the existence in human thought of a double-layered structure, a {\it classical logical thought} and a {\it quantum conceptual thought}, the latter being responsible of the above paradoxes and nonclassical effects. The presence of a quantum structure in cognition is relevant, for it shows that quantum mechanics provides not only a useful modeling tool for experimental data but also supplies a structural model for human and artificial thought processes. This approach has strong connections with theories formalizing meaning, such as semantic analysis, and has also a deep impact on computer science, information retrieval and artificial intelligence. More specifically, the links with information retrieval are discussed in this paper.  
\end{abstract}

\vspace{-0.2 cm}
\section{Introduction}
The aim of this article is to present to the community of quantum technology researchers a new emerging approach to the modeling of human and artificial thought processes that makes use of the formalism of quantum mechanics. We will introduce and describe the main aspects of this new approach which is in full development actually, but has not yet been used for applications of a technological nature. At the same time we also want to specify the specific contributions that the authors, and their collaborators, have made to this new approach in using quantum mechanical techniques and insights for modeling thought processes. There has meanwhile grown a community of researchers, physicists, computer scientists and psychologists, focusing on this approach, and the emerging domain of research has been called `Quantum Cognition'\footnote{We stress that in this paper we are mainly concerned with the Brussels approach to Quantum Cognition. Meanwhile, several researchers have been involved in this domain, and we refer to the Wikipedia page \cite{quantumcognition} for a more complete overview.}. In a general way it studies the application of the formalism of quantum mechanics to situations that traditionally are a subject of investigation in cognitive science, artificial intelligence and semantic theories. More specifically, it aims at modeling entities and/or processes of a cognitive nature, and also put forward potential solutions to some difficult problems in cognitive science, artificial intelligence, semantic theories and information retrieval. Some interesting technological applications of our perspective to symbolic artificial intelligence and robotics will be studied in a forthcoming paper \cite{aertsczachorsozzo2011}. We instead focus here on the applications in representation of knowledge and with respect to information retrieval.

Let us briefly summarize this paper. In Sec. \ref{nonclassicaleffects} we report the problems and paradoxes discovered in decision theory and cognitive science, such as the {\it Allais}  and {\it Ellsberg paradoxes}, the {\it disjunction effect} and the {\it conjunction fallacy}, and the {\it Pet-Fish problem}. In particular, we stress in this section that these difficulties required a deep change of perspective in our understanding of the role played by classical logic in human thought. In Sec. \ref{brussels} we instead report the proposal put forward within our quantum cognition approach. More precisely, we point out that our approach introduces a contextual framework for concept combination. Contextuality makes it possible to elaborate quantum models for the above mentioned paradoxes which correctly reproduce the experimental data existing in the cognitive science literature. An explanation of the obtained results can be given by assuming that two differently structured and superposed, {\it classical logical} and {\it quantum conceptual}, modes coexist in human thought and their simultaneous presence is responsible of the failures of classical logic and probability theory to cope with the aforementioned difficulties. Finally, we discuss in Sec. \ref{ir} the fact that our quantum cognition approach supplies an intuitive support to the use of the quantum mechanical formalism in information retrieval, which is thus more firmly founded.

\section{Nonclassical effects in economics and psychology\label{nonclassicaleffects}}
Note that the identification of the effects in cognition which the quantum cognition community now has identified as needing quantum structures to be modeled were first observed by economists. In two seminal papers Maurice Allais \cite{allais1953} in 1953 -- the year DNA was discovered by Watson and Crick -- and Daniel Ellsberg \cite{ellsberg1961} in 1961 observed that experimental situations exist in economics that exhibit an inconsistency with the traditional economic theory of rational choice, namely a violation of the so-called {\it Sure-Thing Principle} \cite{savage1954} and {\it expected utility hypothesis} \cite{vonneumannmorgenstern1944}\footnote{Ellsberg was studying `decision making' from the White House with respect to the war in Vietnam, having access to all highly classified material with respect to the Vietnam war, when he wrote his seminal article, and it is the same Ellsberg who later released the Pentagon papers containing this classified material, in this way starting the end of the Vietnam war (and risking to go to prison).}. These deviations, often called paradoxes, were at that time identified as indicating the existence of an {\it ambiguity aversion}, that is, individuals prefer `sure choices' over `choices that contain ambiguity' and, eventually `risk taking'. The fact that they were also non-explained deviations from classical logic and probability theory -- which is why solutions of them can be proposed within the new field of quantum cognition -- was not (explicitly) identified then.

The second identification of effects of a non-classical nature was in psychology, and Amos Tversky and Daniel Khaneman played a crucial role in them\footnote{Daniel Khaneman was later to receive the Nobel Price for Economics as a psychologist, and is also considered to be one of the founding fathers of the research domain called `Behavioral Economics', where some of the effects mentioned in the present article are studied.}. The effects brought to light by them have meanwhile been studied extensively in all kinds of ways, and are referred to as the {\it conjunction fallacy} \cite{tverskykahneman1983} and the {\it disjunction effect} \cite{tverskyshafir1992}. Again however, the fundamental contradiction of these effects with classical logic and probability theory, although identified this time, was not believed to be the core of the problem. 

The domain where most manifestly the classical set-theoretical based structures were identified to be failing was in the study of `how concepts combine'. This failure was explicitly revealed by James Hampton's experiments \cite{hampton88a,hampton88b} which measured the deviation from classical set-theoretic membership weights of exemplars with respect to pairs of concepts and their conjunction or disjunction. Hampton's investigation was motivated by the so-called {\it Guppy effect} in concept conjunction found by Osherson and Smith \cite{oshersonsmith81}. These authors considered the concepts {\it Pet} and {\it Fish} and their conjunction {\it Pet-Fish}, and observed that, while an exemplar such as {\it Guppy} was a very typical example of {\it Pet-Fish}, it was neither a very typical example of {\it Pet} nor of {\it Fish}. Therefore, the typicality of a specific exemplar with respect to the conjunction of concepts can show an unexpected behavior from the point of view of classical set and probability theory. As a result of the work of Osherson and Smith, the problem is often referred to as the {\it Pet-Fish problem} and the effect is usually called the {\it Guppy effect}. It can be shown that neither fuzzy set based theories \cite{zadeh01,zadeh02,oshersonsmith02} nor explanation based theories \cite{komatsu01,fodor01,rips03} can model this `typicality effect'. Hampton identified a Guppy-like effect for the membership weights of exemplars with respect to pairs of concepts and their conjunction \cite{hampton88a}, and equally so for the membership weights of exemplars with respect to pairs of concepts and their disjunction \cite{hampton88b}, e.g., {\it Olive} is found to be a not very strong member of {\it Fruits} and also a not very strong member of {\it Vegetables}, but it is a very strong member of {\it Fruits `or' Vegetables}. Hampton called {\it overextension} and {\it underextension} the empirical deviations from the expectations of fuzzy set theory. Several experiments have since been conducted (see, e.g., \cite{hampton01}) and many approaches have been propounded to solve the Pet-Fish problem  and to provide a satisfactory mathematical model of concept combinations, but none of the currently existing concepts theories provides a satisfactory description or explanation of such effects. The combination problem is considered so serious that it is sometimes mentioned that not much progress is possible in the field if no light is shed on this problem \cite{fodor01,rips03,hampton01,kamppartee01}.

Each of the previous problems share a common feature: they have to do with modeling meaning and the structure of human thought, hence a solution to these problems could shed a new light on some disciplines that try to reproduce what happens in human mind, 
such as artificial intelligence and robotics (see \cite{aertsczachorsozzo2011}).   

\section{Quantum contextual structure in human thought\label{brussels}}
Hypothesizing the presence of non-classical logical and probabilistic structures in human thought one of us proposed a quantum model for the modeling of a decision process during an opinion poll \cite{aertsaerts1994}. The inspiration for this proposal came primarily from an extensive study carried out on the nature of the quantum mechanical probability model \cite{aerts1986,aerts1994}, and the becoming aware of the exact and detailed difference between classical probability and quantum probability. 

This perspective suggested that classical probabilistic structures model only situations that are deterministic in essence, and where the observers `lack knowledge about'. Situations of this kind may occur also in quantum mechanics, but this theory predicts the existence of situations that are indeterministic in depth, in the sense they do not admit an underlying deterministic process. For example, in a measurement process the result of the measurement cannot be attributed to the physical entity that is measured, hence the probability of that result in a given state cannot be interpreted as formalizing a subjective lack of knowledge about the entity. It is the interaction of the measured entity with the measurement context that determines the result of the measurement. As a consequence, the measurement context provokes an indeterministic influence on the physical entity, and quantum probability models this indeterminism. 

Whenever the above reasoning is applied to decision processes, it can be shown that models of decision making are quantum in essence, because opinions are not always determined `before the testing of these opinions take place', and can hence in principle easily been influenced by the testing itself \cite{aertsaerts1994}. This is evident if one considers the simple and common situation of an opinion poll, that is, a testing of the dynamics of human decision making with statistics. For certain questions, for example a question like the following one `Are you a smoker or not', the situation for classical probability is satisfied. Indeed, people being tested 'are always smokers or not smokers before the test being applied during the opinion poll'. Hence the test, i.e. poll, for such type of properties `only consists of lifting a lack of knowledge'. However, if one considers another type of question, e.g., `Are you against or in favor of the use of nuclear energy', then it seems to be a better hypothesis about the reality of the states of mind of human beings that there are three cases now (i) the persons mind is made up already, in favor; (ii) the persons mind if made up already, against; (these are the two `classical probability situations, the test only lifts lack of knowledge'); (iii) the persons mind is not made up, and formed `during and hence partly by the test itself'. It is this third possibility which introduced a `non classical probabilistic effect'. As a consequence, human decision making is intrinsically nonclassical, and quantum probability is an obvious choice as an alternative to classical probability, since in quantum probability this extra aspect is `exactly the one which is allowed to exist'. Namely, entities being tested upon may be in states that are undetermined in a sense different from the `lack of knowledge situation' (which is the classical one).

The formalism ({\it SCoP formalism}) was developed to cope with the difficulties presented in Sec. \ref{nonclassicaleffects}. It is a generalization of the quantum formalism and in which context plays a relevant role in both concept representation and decision processes \cite{gaboraaerts2002,aertsgabora2005a,aertsgabora2005b,aertsczachordhooghe2006}. This role is very similar to the role played by the measurement context on microscopic entities in quantum mechanics. In the SCoP formalism each concept is associated with well defined sets of states, contexts and properties (see Appendix A). Concepts change continuously under the influence of a context and this change is described by a change of the state of the concept. For each exemplar of a concept, the typicality varies with respect to the context that influences it. Analogously, for each property, the applicability varies with respect to the context. This implies the presence of both a {\it contextual typicality} and an {\it applicability effect}. The {\it Pet-Fish problem} is solved in the SCoP formalism because in different combinations the concepts are in different states. In particular, in the combination {\it Pet-Fish} the concept {\it Pet} is in a state under the context {\it The Pet is a Fish}. The state of {\it Pet} under the context {\it The Pet is a Fish} has different typicalities, which explains the guppy effect. On the basis of the SCoP formalism, a mathematical model using the formalism of quantum mechanics, both the quantum probability and Hilbert space models, has been worked out which allows one to reproduce the experimental results obtained by Hampton on conjunctions and disjunctions of concepts. This formulation identifies the presence of typically quantum effects in the mechanism of combination of concepts, e.g., contextual influence, superposition, interference, emergence and entanglement \cite{aerts2009a,aerts2009c,aertsaertsgabora2009,aertsdhooghe2009,aerts2007a,aerts2007b} (see Appendix B). Furthermore, quantum models have also been elaborated to describe the disjunction effect and the Ellsberg paradox, which accord with the experimental data existing in the literature \cite{aerts2009c,aertsdhooghe2009,aertsdhooghehaven2010}. It has been shown that it is the overall conceptual landscape, or context, that generates the paradoxical situation encountered in real experiments. 

The analysis above has allowed the authors to suggest the hypothesis that two structured and superposed layers can be identified in human thought: a {\it classical logical layer}, that can be modeled by using a classical Kolmogorovian probability framework, and a {\it quantum conceptual layer}, that can instead be modeled by using the (nonKolmogorovian) probabilistic formalism of quantum mechanics. The thought process in the latter layer is given form under the influence of the totality of the surrounding conceptual landscape, hence context effects are fundamental in this layer.

We conclude this section with two remarks. Firstly, we note that in our approach the explanation of the violation of the expected utility hypothesis and the Sure-Thing Principle is not (only) the presence of an ambiguity aversion. On the contrary, we argue that the above violation is due to the concurrence of superposed conceptual landscapes in human minds, of which some might be linked to ambiguity aversion, but other completely not. We therefore maintain that the violation of the Sure-Thing Principle should not be considered as a fallacy of human thought, as often claimed in the literature but, rather, as the proof that real subjects follow a different way of thinking than the one dictated by classical logic in some specific situations, which is context-dependent. Secondly, we observe that according to our approach `there is a definite holistic aspect' related to the structuring of meaning. However, this holistic aspect is not the one mentioned sometimes with reference to `quantum consciousness' or other very speculative related issues, the presence of holism is of a `down to earth' nature and simply revealing that `meaning is related to the whole landscape of conceptual, emotional, \ldots, content'.

\section{Quantum cognition and information retrieval\label{ir}}
Quantum mechanical structures have also been used in the domain of information retrieval \cite{widdows2003,widdows2006,widdowspeters2003,vanrijsbergen2004}. An information retrieval system finds relevant information from a collection of information objects, which may be documents, web pages, images, videos, etc. For example, search engine algorithms exploit the link structure of the web besides the lexical content in web pages. Roughly speaking, they mine human decisions about what pages are good to link to for a particular subject. Novel techniques developed since the eighties have shown that vector space based information retrieval is potentially more powerful than Boolean logic based information retrieval (in particular, key words matching). In recent works vector space models have been extended to incorporate Hilbert spaces together with the representation and manipulation of word meaning. It has been shown, in particular, that it is the quantum logic structure underlying quantum mechanics that has proved effective in producing theoretical models for information retrieval. Widdows and Peters \cite{widdowspeters2003} proved that the use of quantum logic connective for negation, i.e. orthogonality, is a powerful tool for exploring and analyzing word meaning and it is more efficient than Boolean `not' of classical logic in document retrieval experiments. Van Rijsbergen \cite{vanrijsbergen2004} introduced a general theory for information retrieval based on quantum logic structures. More precisely, he showed that three keystone models used in information retrieval, a vector space model, a probabilistic model and a logical model, can be described within Hilbert space, where a document can be represented by a vector and relevance by a Hermitian operator. 

The above results fit in naturally with our quantum cognition approach resumed in Sec. \ref{brussels}, which provides theoretical support for the use of the quantum mechanical formalism in information retrieval and natural language processing in terms of the quantum conceptual layer. It is quite reasonable, indeed, to maintain that a given human or artificial system aiming at extracting information and knowledge from a user should be quantum based, since concepts are combined by human minds in such a way that they entail quantum mechanical structure, more specifically for example, combined concepts are entangled \cite{aerts2009c,aertssozzo2011}. But there are other strong reasons for claiming the necessity of using contextual quantum based structures in information retrieval, and also this arises from the research developed in our quantum cognition approach. In fact, the following results have been obtained by using search engines on the world-wide-web.

(i) A Guppy-like effect can be identified by collecting data on concepts and their conjunctions by means of web search engines. This effect appears as a consequence of the contextual meaning landscape surrounding the concepts considered and their conjunctions \cite{aertsczachordhooghesozzo2010}. 

(ii) Some Bell's inequalities can be deduced by considering coincidence experiments and gathering data on concepts and their combinations on the world-wide-web. One can show that these inequalities are violated, and this violation reveals the presence of a typically quantum effect, namely, entanglement \cite{aerts2010}.

Let us conclude this paper by pointing out a possible research line. Knowledge organization systems (KOS) play an increasingly important role in modern society \cite{hodge2000,stock2010}. They employ classical structures for the properties and concepts used in their knowledge models and hence are subject to the difficulties mentioned in Sec. \ref{nonclassicaleffects}. But, one could provide a generalized, or `quantized', version of a KOS by using our SCoP quantum based  formalism for the properties and concepts (see Appendix A). Taking into account the foregoing analysis, together with the results obtained by van Rijsbergen and Widdows, we expect that a SCOP version of KOS will give rise to greater adaptiveness and performance. 

\section*{Appendix A\label{A}}
In the SCoP formalism a concept is an entity that can be in different states, and such that a given context provokes a change of state of the concept. Let $S$ be a concept. We denote by $\Sigma$ the set of its states, by ${\cal M}$ the set of its contexts and by ${\cal L}$ the set of its properties. Moreover, we introduce a special state of the concept $S$ called the {\it ground state}, denoting it by $\hat{p}$. The ground state can be considered as the state the concept is in when it is not triggered by any particular context. Furthermore, we introduce for a given concept $S$ a {\it unit context} 1 which describes the situation where no context occurs. A given context $e_1$ induces a change of state of the concept $S$ from the the ground state $\hat{p}$ to another state, say $p_1$. That $\hat{p}$ and $p_1$ are different states is manifested by the fact that the frequency measures of different exemplars of the concept, as well as the applicability values of the properties of the context are different for different states.

To elaborate a SCoP system we do not only need of the three sets $\Sigma, {\cal M}$ and ${\cal L}$, that is, the set of states, the set of contexts and the set of
properties, respectively, but also two additional functions $\mu$ and $\nu$ are required. The function $\mu$ is a  probability function that describes how state $p$ under the influence of context $e$ changes to state $q$. Mathematically, this means that $\mu$ is a function from the set $\Sigma \times {\cal M} \times \Sigma $ to the interval $[0, 1]$, where $\mu(q, e, p)$ is the probability that state $p$ under the influence of context $e$ changes to state $q$. We write $\mu: \Sigma \times {\cal M} \times \Sigma \rightarrow [0, 1]; (q, e, p) \mapsto \mu(q, e, p)$. The function $\nu$ describes the weight (the renormalization of the applicability) of a certain property given a specific state. This means that $\nu$ is a function from the set $\Sigma \times {\cal L}$ to the interval $[0, 1]$, where $\nu(p, a)$ is the weight of property $a$ for the concept in state $p$. We write $\nu: \Sigma \times {\cal L} \rightarrow [0, 1]; (p, a) \mapsto \nu(p, a)$. Thus a SCoP system is defined by the five elements $(\Sigma, {\cal M}, {\cal L}, \mu, \nu)$. Up until this point, the SCoP system we have built for the concept `pet' has been rather small. To work out a more elaborate SCoP system, we proceed as follows. We collect all the contexts thought to be relevant to the model we want to build (more contexts lead to a more refined model). ${\cal M}$ is the set of these contexts. Starting from the ground state $\hat{p}$ for the concept, we collect all the new states of the concept formed by having each context $e \in {\cal M}$ work on $\hat{p}$ and consecutively on all the other states. This gives the set $\Sigma$. Note that ${\cal M}$ and $\Sigma$ are connected in the sense that to complete the model it is necessary to consider the effect of each context on each state. We collect the set of relevant properties of the concept and this gives ${\cal L}$. The functions $\mu$ and $\nu$ that define the metric structure of the SCoP system have to be determined by means of well chosen experiments. 

\section*{Appendix B}
It is interesting to observe that a typically quantum effect, i.e. an interference effect, in concept combination appears at once if one considers the following simple quantum model \cite{aerts2009a}. 

Let us consider the two concepts $A$ and $B$. Both $A$ and $B$ are described quantum mechanically in a Hilbert space ${\cal H}$, so that they are represented by the unit vectors $|A\rangle$ and $|B\rangle$ of ${\cal H}$, respectively. We describe concept `$A$ or $B$' by means of the normalized superposition vector ${1 \over \sqrt{2}}(|A\rangle+|B\rangle)$, and also suppose that $|A\rangle$ and $|B\rangle$ are orthogonal, hence $\langle A|B\rangle=0$. An experiment considered in Hampton \cite{hampton88a,hampton88b} consists in a test aimed to ascertain whether a specific item $X$ is `a member of' or `not a member of' a concept. We represent this experiment by means of a projection operator $M$ on this Hilbert space ${\cal H}$. This experiment is applied to concept $A$, to concept $B$, and to concept `$A$ or $B$', respectively, yielding specific probabilities $\mu(A)$, $\mu(B)$ and $\mu(A\ {\rm or}\ B)$. These probabilities represent the degrees to which a subject is likely to choose $X$ to be a member of $A$, $B$ and `$A$ or $B$'. In accordance with the quantum rules, these probabilities are given by $\mu(A)=\langle A|M|A\rangle \nonumber$, $\mu(B)=\langle B|M|B\rangle \nonumber$ and
\begin{displaymath}
 \mu(A\ {\rm or}\ B)={1 \over 2}(\langle A|+\langle B|)M(|A\rangle+|B\rangle) .
\end{displaymath}
Applying the linearity of Hilbert space and taking into account that $\langle B|M|A\rangle^*=\langle A|M|B\rangle$, we have
\begin{eqnarray}
\mu(A\ {\rm or}\ B)={1 \over 2}(\langle A|M|A\rangle+ \\ \nonumber 
\langle A|M|B\rangle+ \langle B|M|A\rangle+\langle B|M|B\rangle)=  \\ \nonumber
{\mu(A)+\mu(B) \over 2}+\Re\langle A|M|B\rangle \nonumber
\end{eqnarray}
where $\Re\langle A|M|B\rangle$ is the real part of the complex number $\langle A|M|B\rangle$. This is called the `interference term' in quantum mechanics. Its presence produces a deviation from the average value ${1 \over 2}(\mu(A)+\mu(B))$, which would be the outcome in the absence of interference. It is the interference term that is responsible of the deviations from classically expected membership weights measured by Hampton. The above quantum mechanical formalism in Hilbert space has been used in \cite{aerts2009c} to model all the experimental data obtained by James Hampton for experiments published in \cite{hampton88a,hampton88b}.

\section*{Acknowledgment}
This research was supported by Grants G.0405.08 and G.0234.08 of the Flemish Fund for Scientific Research.

\end{document}